\documentclass[conference]{IEEEtran}
\IEEEoverridecommandlockouts
\usepackage{cite}
\usepackage{amsmath,amssymb,amsfonts}
\usepackage{algorithmic}
\usepackage{graphicx}
\usepackage{textcomp}
\usepackage{xcolor}

\usepackage{multirow}
\usepackage{bbding}
\usepackage{bm}
\usepackage{pifont}
\usepackage{fontawesome}
\usepackage{colortbl}
\usepackage{array}
\usepackage{makecell}
\definecolor{Gray}{gray}{0.9}
\definecolor{LightCyan}{rgb}{0.88,0.95,1}
\usepackage{booktabs}

\def\BibTeX{{\rm B\kern-.05em{\sc i\kern-.025em b}\kern-.08em
    T\kern-.1667em\lower.7ex\hbox{E}\kern-.125emX}}
\begin{document}

\title{MuseFace: Text-driven Face Editing via Diffusion-based Mask Generation Approach}

\author{
    \IEEEauthorblockN{
            Xin Zhang\IEEEauthorrefmark{2}$^{,}$\IEEEauthorrefmark{3}, 
            Siting Huang\IEEEauthorrefmark{3},
            Xiangyang Luo\IEEEauthorrefmark{2},
            Yifan Xie\IEEEauthorrefmark{2}$^{,}$\IEEEauthorrefmark{3}, 
            Weijiang Yu\IEEEauthorrefmark{4}, 
            Heng Chang\IEEEauthorrefmark{5}, 
            Fei Ma\IEEEauthorrefmark{2}$^{,}$\IEEEauthorrefmark{7} and Fei Yu\IEEEauthorrefmark{2}
        }
    \vspace{0.1cm}
    \IEEEauthorblockA{
        \IEEEauthorrefmark{2}Guangdong Laboratory of Artificial Intelligence and Digital Economy (SZ), Shenzhen, China \\
    }
    \IEEEauthorblockA{
        \IEEEauthorrefmark{3}Xi’an Jiaotong University, Xi'an, China \\
    }
    \IEEEauthorblockA{
        \IEEEauthorrefmark{4}Sun Yat-sen University, Guangzhou, China \\
    }
    \IEEEauthorblockA{
        \IEEEauthorrefmark{5}Tsinghua University, Shenzhen, China \\
    \vspace{0.1cm}
    Email: mafei@gml.ac.cn
    }
    \vspace{-6mm}
    % \thanks{This work was done during Xin’s Intern at Guangdong Laboratory of Artificial Intelligence and Digital Economy (SZ).}
    \thanks{$^{**}$Corresponding author.}
}

\maketitle

\begin{abstract}
Face editing modifies the appearance of face, which plays a key role in customization and enhancement of personal images. Although much work have achieved remarkable success in text-driven face editing, they still face significant challenges as none of them simultaneously fulfill the characteristics of diversity, controllability and flexibility. To address this challenge, we propose MuseFace, a text-driven face editing framework, which relies solely on text prompt to enable face editing. Specifically, MuseFace integrates a Text-to-Mask diffusion model and a semantic-aware face editing model, capable of directly generating fine-grained semantic masks from text and performing face editing. The Text-to-Mask diffusion model provides \textit{diversity} and \textit{flexibility} to the framework, while the semantic-aware face editing model ensures \textit{controllability} of the framework. Our framework can create fine-grained semantic masks, making precise face editing possible, and significantly enhancing the controllability and flexibility of face editing models. Extensive experiments demonstrate that MuseFace achieves superior high-fidelity performance.
\end{abstract}


\section{Introduction}
In the rapidly evolving field of Artificial Intelligence Generated Content (AIGC)~\cite{karras2019style, xie2024pointtalkaudiodrivendynamiclip}, image manipulation~\cite{10126081, Luo2024Code} emerges as a pivotal technique, bridging the gap between reality and the boundless potential of visual creativity. Among the various applications, face editing emerges as particularly prominent, offering substantial promise. This technique concentrates on modifying specific local regions of a face image, guided by textual descriptions or mask conditions. Previous efforts have explored this task with current state-of-the-art generative methods such as Generating Adversarial Networks (GAN)~\cite{goodfellow2020generative}, Variational Autoencoders (VAE)~\cite{kingma2013auto} and large-scale diffusion model~\cite{rombach2022high, ramesh2022hierarchical, saharia2022photorealistic, ho2021classifier}.

% Recently, major advancements in large-scale text-to-image diffusion model~\cite{rombach2022high, ramesh2022hierarchical, saharia2022photorealistic, ho2021classifier} have demonstrated impressive capabilities in image generation with diverse prompts.

However, there still remain significant challenges, with a critical problem yet to be resolved: how to generate \textit{diverse} editing results in a \textit{controllable} and \textit{flexible} manner. Initially, some of the work~\cite{zhu2017unpaired,durall2021facialgan} transfers some of the face attributes from the reference image to the target image in the form of style transfer. However, the completion of the editing is completely limited by the number and variety of reference images and lacked creative generation. Subsequently, the integration of textual input facilitates more creative and diverse editing outcomes from the model. Regrettably, the model may struggle to understand the contextual relationship between the unmasked region and the reference text prompt, leading to artifacts and visually unreasonable completions~\cite{jiang2021talk, yue2023chatface}. To address these issues, TediGAN~\cite{xia2021tedigan} and CollabDiff~\cite{huang2023collaborative} improve performance by using text and semantic masks as inputs to face editing. Nevertheless, the labor-intensive process of generating fine-grained masks contrasts with the inherent limitations of coarse-grained alternatives such as bounding boxes, significantly impeding progress in multimodal face editing tasks. To sum up, existing methods have contributed to the field of face editing, yet none of them simultaneously fulfill the three characteristics of \textit{controllability}, \textit{diversity}, and \textit{flexibility}.

Motivated by these critical gaps, we introduce a text-driven face editing framework named MuseFace. Firstly, the core idea of MuseFace is to generate fine-grained masks that are driven by text alone, then guide face editing. To achieve this, we design a novel diffusion model capable of editing face semantic masks based on text guidance, with the \textit{flexibility} of text input for \textit{diverse} generation of face semantic masks. Subsequently, we proprose a semantic-aware face editing model to perform face image editing with semantic masks as conditions, which provides a more controlled editing prior. 

\begin{figure}
 % \vskip -0.15in 
  %\setlength{\abovecaptionskip}{0.2cm}
  \includegraphics[width=0.5\textwidth]{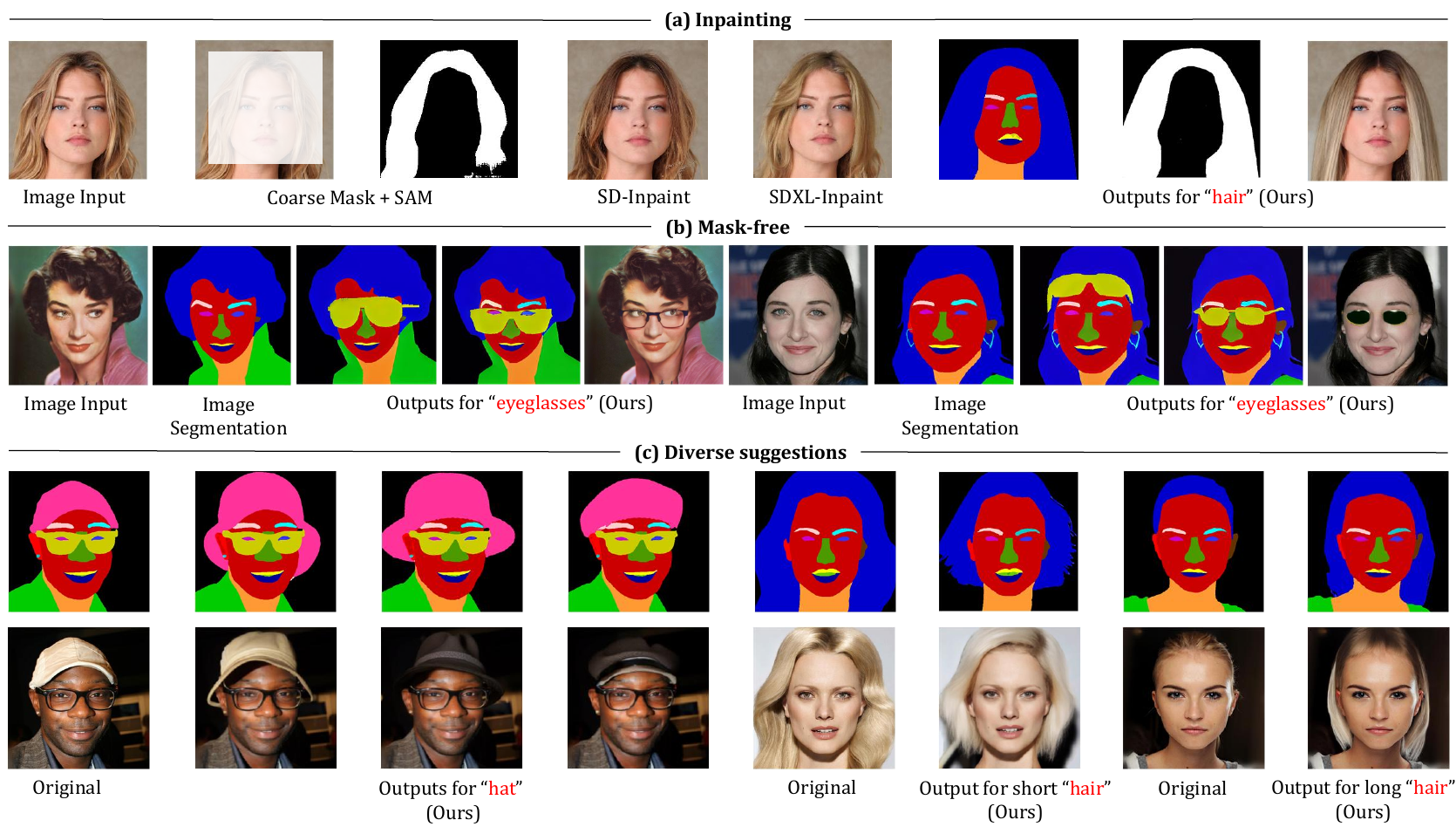}
  \caption{\textbf{Overview}. We introduce MuseFace which allows users to generate high-fidelity face editing results. \textbf{(a)} MuseFace first generate semantic map where the shape has been edited, then edit the reference image guided by the mask (e.g., hair). \textbf{(b)} and \textbf{(c)} MuseFace is capable of generating diverse outputs with greater controllability based solely on text prompt (flexibility). }
  % \Description{Enjoying the baseball game from the third-base
  % seats. Ichiro Suzuki preparing to bat.}
  \label{fig:teaser}
  %\vskip 0.1in
\end{figure}

In particular, given a reference image $\mathcal{I}$ and a text prompt $\mathcal{T}_{edit}$ describing a specified edit area, MuseFace initially generates diverse location-aware fine-grained mask outputs. Subsequently, these mask outputs are employed within the semantic-aware face editing model to facilitate editing effects of high fidelity (Fig. \ref{fig:teaser} (c)). This approach enhances the controllability of face editing by introducing greater diversity and flexibility (Fig. \ref{fig:teaser} (b)). Additionally, MuseFace provides a highly convenient and controllable method for face editing compared with coarse-mask-based inpainting methods (Fig.~\ref{fig:teaser} (a)). This is because our method can be utilized in two distinct manners: 1) in conjunction with user-supplied coarse masks (such as bounding boxes), allowing users to define the position and configuration of the target object, or 2) through a mask-free modality, wherein the model autonomously generates a variety of suggestions across different locations and scales. 

% To achieve this, we design a novel diffusion model capable of editing face semantic maps based on text guidance. Besides, we take advantage of a learnable skip connection that transfers the details from the encoding phase to the corresponding decoding one to extend the autoencoder architecture, namely Mask-aware Autoencoder. Mask-aware Autoencoder improves the accuracy of semantic map reconstruction, resulting in high quality of mask generation. We also train a multimodal face editing model to perform face image editing as a downstream of the framework.

In summary, the main contributions of the paper are: 
\begin{itemize}
\item We propose a text-driven framework for high-fidelity face editing named MuseFace, which consists of a Text-to-Mask diffusion model and a semantic-aware face editing with simultaneous controllability, diversity, and flexibility.

% \item We introduce a diffusion-based Text-to-Mask diffusion model with a Mask-aware Autoencoder,  which allows any user to generate precise fine-grained masks of face for the first time. Moerover, we also train a multimodal face editing model relying on the mask data generated by Text-to-Mask diffusion model.
\item We introduce a novel diffusion-based Text-to-Mask face editing model, enabling users to generate precise fine-grained masks of face for text-driven face editing. Moreover, a semantic-aware face editing model is proposed to perform downstream task.
% In MuseFace, We first design diffusion-based Text2Mask which allows any user to generate precise fine-grained masks of face for multimodal editing and object insertion with other attributes shape preservation. Further, relying on the data generated by Text2Mask, we trained a ControlNet\cite{zhang2023adding} model to perform high-fidelity multimodal face editing.
% \item We employ a Mask-aware Autoencoder that accurately reconstructs semantic label of each pixel, which is designed to pass information from the encoding process to the corresponding decoding process through a design skip connection module. This is a plug-and-play module that can be adapted to other stable diffusion models.
% \item We extensively validate the effectiveness and zero-shot capability of MuseFace on human face datasets.  We believe our results can highlight how fine-grained masks can strongly benefit from using LDMs and serve as a starting point for future research of multimodal face editing in the field.
\item  Extensive experiments demonstrate the effectiveness and zero-shot capability of the proposed framework.
% \item Based on the CelebAMask-HQ datasets, we can extend the datasets by generating a multi-modal training dataset containing text editing instructions and the corresponding mask before and after the edit.
\end{itemize}

% In the following, Section \ref{related work} gives a brief review of face editing tasks and mask generation. Section \ref{method} details the proposed framework. Section \ref{experiments} provides the experimental settings, evaluation results. Section \ref{discussion} further develops the discussion. Finally, Section \ref{conclusion} draws the conclusion and presents potential future works.

\section{Method}\label{method}

\subsection{Motivation}
\begin{figure}[h]
  \centering
  \includegraphics[width=0.5\textwidth]{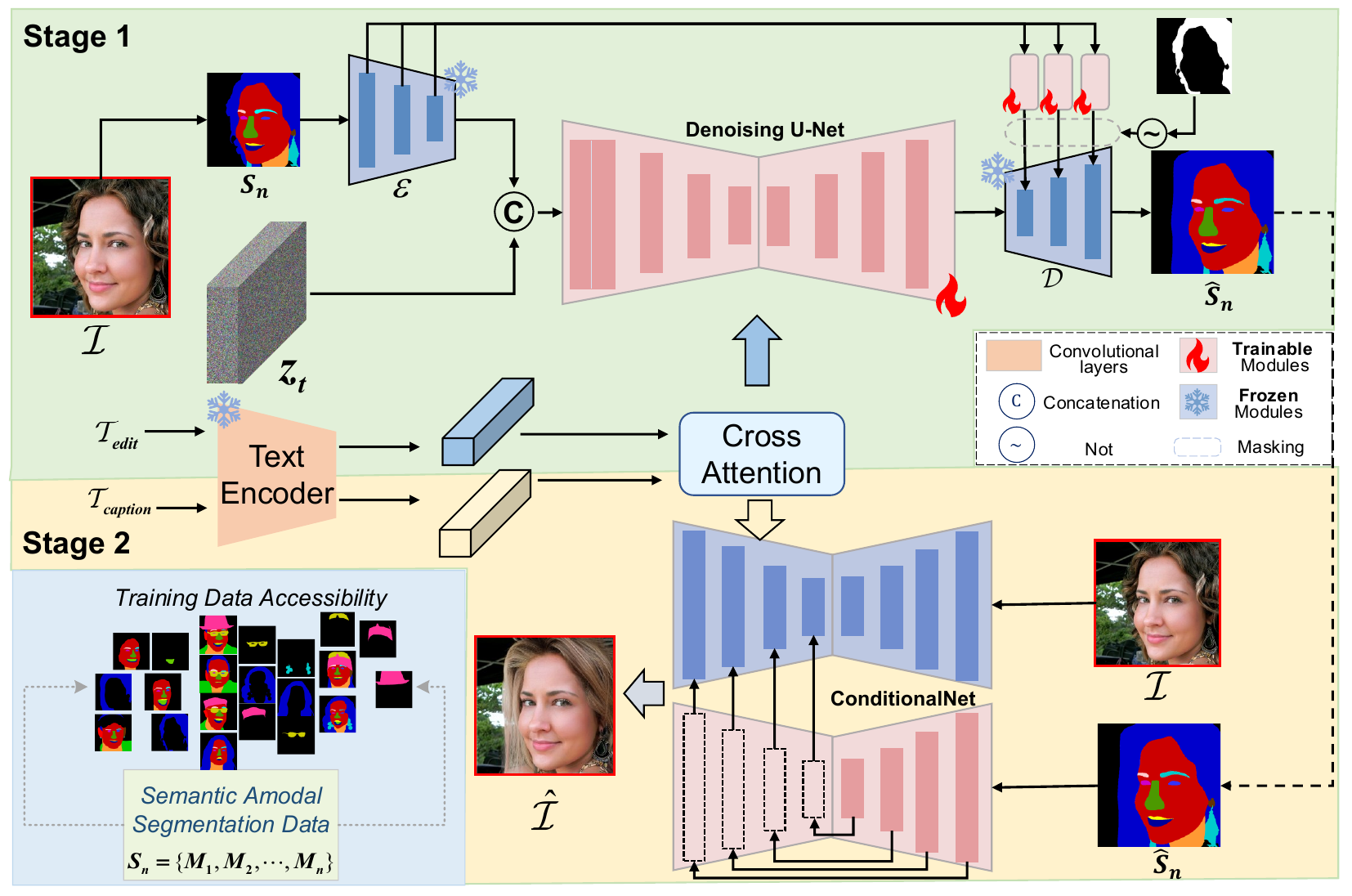}
  \caption{The overall pipeline of our proposed MuseFace which consists of two diffusion models, the Text-to-Mask diffusion model and the semantic-aware face editing model. The inputs of MuseFace are only reference images $\mathcal{I}$ text $\mathcal{T}_{edit}$ specifying the part to be edited, and the caption $ \mathcal{T}_{caption}$ of $\mathcal{\hat{I}}$. The Text-to-Mask diffusion model edits the semantic map driven by $\mathcal{T}_{edit}$ and the output is used for face editing model to generate the edit face image. The dashed box in the lower left corner is a schematic of the training data accessibility. }
  % \Description{A woman and a girl in white dresses sit in an open car.}
  \label{fig:overflow model}
  % \vskip -0.1in
\end{figure}

First, from a data-driven perspective, learning a model directly in pixel space for editing tasks presents significant challenges, as it requires extensive collection of paired training data to ensure preservation of unintended features~\cite{singh2023smartmask}. Instructpix2pix~\cite{brooks2023instructpix2pix} attempts to generate a large number of paired images to train a diffusion model for editing. However, the model editing results will be less realistic by training with generated data. Therefore, our creative solution is to transform pixel space editing tasks into semantic space editing tasks, which addresses the challenge of insufficient large-scale pairwise training data and enhancing the controllability of editing. Then, from a model design perspective, a Text-to-Mask diffusion model is proposed with a series of innovative adaptations to effectively manipulate semantic masks (Sec. \ref{sec:Text-to-Mask diffusion Model}). Finally, from a task accomplishment perspective, the generated semantic masks are then utilized to guide the natural image editing process with proposed semantic-aware face editing model, completing the final face editing task (Sec. \ref{sec: Semantic-aware Face Editing Model}). Fig. \ref{fig:overflow model} depicts an overview of the proposed MuseFace, consisting of proposed Text-to-Mask diffusion model and semantic-aware face editing model (Sec. \ref{sec:infer}).

\subsection{Stage 1: Text-to-Mask Diffusion Model}
\label{sec:Text-to-Mask diffusion Model}
Given an input face image $\mathcal{I}$ and description of object to be edited $\mathcal{T}_{edit}$, our model can generate a fine-grained semantic mask $\hat{S}_{n}$ where the part described by $\mathcal{T}_{edit}$ is edited from the original shape to a new shape guided by $\mathcal{T}_{edit}$ with the rest parts preservation.

To achieve this, we leverage large-scale semantic amodal segmentation data~\cite{qi2019amodal} to create high quality paired training data in the semantic space. This approach is relatively simpler and can serve as a foundation for other tasks. In particular, given a human face image $I$, with a sequence of amodal semantic instance maps ${M_1, M_2, \dots, M_n}$ and corresponding semantic object labels ${t_1,t_2,\dots,t_n}$, we randomly choose $k \in [1,n)$ and compute an intermediate semantic map as:

\begin{equation}
\begin{aligned}
 S_k = f_{layer}(\{M_1, M_2, \dots, M_{k-1}, M_{k+1}, \dots, M_n\}),
 \label{eq:intermediate-layer}
 \end{aligned}
\end{equation}
where $f_{layer}$ is an operation which stacks the semantic instance maps and $k=n$ is the semantic segmentation map. Therefore, we edit $M_k$ of $S_n$ to a new shape when editing semantic object $t_k$ of image $\mathcal{I}$. We prepare paired data for each semantic object label $t_k$ for training.

As for model architecture, the Text-to-Mask diffusion model consists of a Mask-aware Autoencoder and a denoising U-Net for diffusion face mask generation. Firstly, An autoencoder which learns a space that is perceptually equivalent to the semantic space is proposed. However, directly compressing and reconstructing the face semantic segmentation map with the autoencoder results in noticeable discrepancies at the pixel level, significantly affecting the quality of the semantic segmentation map. To address the problem, we extend the Autoencoder with an additional branch that encodes the intermediate semantic map $S_k$ and sums each layer of features in the encoder to the corresponding layer of the decoder in the main branch via a learnable skip connetction module~\cite{morelli2023ladi}. To facilitate this, mask-aware information is passed from the distinct layers of the encoder $\mathcal{E}$ to the matching layers of the decoder $\mathcal{D}$. 
This process allows features that are not altered during the editing task to filter through, keeping the editing process shape-agnostic at the initial site of modification. Additionally, the intermediate features of the encoded image $S_k$ are fed into a learnable convolutional module $\Phi_\alpha$ and then skip-connected to the decoder for reconstruction, enabling the Autoencoder to adaptively perceive mask information, namely Mask-aware. This design also allows the user to provide coarse-grained mask for more precise editorial control. 
We formulate this process as followed:
\begin{equation}
 \mathcal{D}_i = \mathcal{D}_{i-1} + (\Phi_\alpha(E_i) \odot (\mathcal{J}-\rho_i(M))),
 \label{eq:mask-aware}
\end{equation}
where $i$ denotes the $i$-th $DownEncoderBlock$ in the Encoder $\mathcal{E}$ or $UpDecoderBlock$ in Decoder $\mathcal{D}$ and $\mathcal{J}$ is all-ones matrix. $\rho_i(M)$ resizes the mask input $M$ to match the spatial dimension of the corresponding feature map.

The modules are trained by employing a combination of the L1 and perceptual loss function.
\begin{equation}
 L_\alpha(S_n,\hat{S_n}) = \lambda_1 \cdot \Vert S_n - \hat{S_n} \Vert_1 + \lambda_2 \cdot \Vert \text{VGG}(S_n) - \text{VGG}(\hat{S_n}) \Vert_2,
 \label{eq:autoencoder}
\end{equation}
where $\hat{S_n}$ is reconstructed semantic map, and $\text{VGG}(\cdot)$ denotes the VGG perceptual loss function. Please refer the supp. material the overview of Mask-aware Autoencoder.

% \begin{figure}[h]
%   \centering
%   % \setlength{\abovecaptionskip}{0.cm}
%   \includegraphics[width=0.5\textwidth]{./figure/mask-aware.pdf}
%   \caption{Overview of the proposed autoencoder with Mask-aware modules.}
%   % \Description{A woman and a girl in white dresses sit in an open car.}
%   \label{fig:autoencoder}
%   % \vskip -0.1in
% \end{figure}

We demonstrate that this module enhances the model's capacity to adapt to semantic masks comprised of pixels that respond to perceptual information, thereby facilitating precise pixel-level reconstruction outcomes. This feature allows editing the semantic masks to modify regions of interest while keeping other areas unchanged, which is essential for localized editing. However, this is not conducive to the task of adding objects. Hence, MuseFace sets the Mask-aware Autoencoder as a plug-and-play module. Please refer the Section \ref{Ablation Study} for ablation study result.

\textbf{Diffusion Face Mask Generation. }
The Mask-aware Autoencoder enhances the reconstruction of the semantic masks, while editing the semantic map is performed by diffusion model. To perform the semantic map editing task, we extend the U-Net denosier $\epsilon_\theta$ of our diffusion-based mask generation model $\mathcal{M}_\theta$ takes as inputs intermediate semantic masks $S_k$, and textual description $T_{edit}$ specifying the part to be edited, conditioning the diffusion pipeline. Thus, our task can be formulated as followed:
\begin{equation}
 \hat{S_n} = \mathcal{M}_\theta(Z_T, S_k, \mathcal{T}_{edit}).
 \label{eq:task formulation}
\end{equation}

Because Latent Diffusion Models~\cite{rombach2022high} take as spatial input $Z_t \in \mathbb{R}^{4 \times h \times w}$, we extend the spatial input concatenating it with $\mathcal{E}(S_k) \in \mathbb{R}^{4 \times h \times w}$. The final spatial input $\gamma=[Z_t;\mathcal{E}(S_k)] \in \mathbb{R}^{(4+4) \times h \times w}$. At the same time, we also extend kernel channels of the first convolutional layer by adding zero initialized weights to match the new input channel dimension and enrich the input capacity of the diffusion model. In such a way, we can retain the knowledge embedded in the original denoising network while allowing the model to deal with the newly proposed inputs~\cite{morelli2023ladi}. 
We next train the U-Net denosier $\epsilon_\theta$ of the diffusion-based mask generation model $\mathcal{M}_\theta$. To achieve this, we first pass the intermediate semantic map $S_k$ through Mask-aware Autoencoder $\mathcal{E}$ to obtain the encoded features $\mathcal{E}(S_k)$. We train the proposed the U-Net denosier $\epsilon_\theta$ to predict the noise stochastically added to previous noise map $Z_t$. Specifically, at any timestep $t$ of the reverse diffusion process, the denoising noise prediction $\epsilon_t$ is then computed conditional jointly on previous noise map $Z_t$ as,
\begin{equation}
 \hat{\epsilon_t} = \epsilon_\theta(Z_t, \mathcal{E}(S_k), \mathcal{T}_{edit},t).
 \label{eq:eps pred}
\end{equation}
Finally, we specify the corresponding objective function as:
\begin{equation}
\mathcal{L}_{t}(\theta) = \mathbf{E}_{t\sim[1,T], S_k, \epsilon_t}[\Vert \mathbf{\epsilon}_t - \mathbf{\hat{\epsilon}}_{t}\Vert^2],
\end{equation}
where $T$ is total number of reverse diffusion steps, and $\mathbf{\epsilon}_t \sim \mathcal{N}(0,I)$.

\subsection{Stage 2: Semantic-aware Face Editing Model}\label{sec: Semantic-aware Face Editing Model}
Conditional diffusion models commonly employ two methods to integrate mask conditions that deliver spatial information, where the mask is either concatenated with the noise as an input to the U-Net~\cite{rombach2022high} or condition the diffusion process through an Adapter~\cite{mou2024t2i}. We argue that this approach fails to fully leverage the benefits of fine-grained masks. Thus, we design a multimodal face editing model $\mathcal{F}$ consisting of a conditional network that encodes complex features in semantic map images with additional spatial control~\cite{zhang2023adding}. Specifically, we first freeze the U-Net which performs the prediction of noise for image generation (not mask generation). We then obtain a trainable copy of this U-Net encoder blocks and middle block. Each block's output is skip-connected to a zero-convolutional layer, and the result is then combined with the corresponding block of the frozen U-Net to condition the image generation process. This trainable network takes the mask input $c_m$ as a condition and its output is added to text-to-image inpainting model, as shown in Fig. \ref{fig:overflow model}. During training, we only optimize the parameters of the conditional network:
\begin{equation}
		\mathcal{L} = \mathbb{E}_{\bm{z}_0, \bm{t}, \bm{c}_t, \bm{c}_\text{m}, \epsilon \sim \mathcal{N}(0, 1) }\Big[ \Vert \epsilon - \epsilon_\theta(\bm{z}_{t}, \bm{t}, \bm{c}_t, \bm{c}_\text{m})) \Vert_{2}^{2}\Big],
		\label{eq:loss}
	\end{equation}
where $\mathcal{L}$ is the overall learning objective of the entire diffusion model, and $\bm{c}_t$ and $\bm{c}_\text{m}$ are the text condition and mask condition respectively. Note that the face editing model and Text-to-Mask are two diffusion models with different U-Net architecture.
% Stable diffusion has shown good performane in image inpainting tasks. Thus, we add spatial condition $C_m$ to Stable diffusion backbone to perform multimodal face editing.
\subsection{MuseFace Inference}\label{sec:infer}
In inference time, given a face image $I$, text prompt of the editing object $\mathcal{T}_{edit}$, and a caption of image $\mathcal{T}_{caption}$, we first use a face parsing model~\cite{lee2020maskgan} to obtain the corresponding ground truth semantic map $S_{n}$ which is then directly used as input to the above trained diffusion model $\mathcal{M}_\theta$ to generate a new semantic map $\hat{S}_{n}$, where editing object has been modified based on the text prompt $\mathcal{T}_{edit}$ while the rest remains unchanged:
\begin{equation}
 \hat{S}_{n} = \mathcal{M}_\theta(Z_t, \mathcal{E}(S_{n}), \mathcal{T}_{edit}).
 \label{generate process}
\end{equation}

Furthermore, the semantic-aware face editing model $\mathcal{F}$ takes as the input the post-edit semantic map $\hat{S}_{n}$ and $T_{caption}$ to edit the reference image $I$. To summarize, Stage 1 is responsible for editing the shapes and Stage 2 completes the editing of the textures.
\begin{equation}
 \hat{I} = \mathcal{F}(I, \hat{S}_{n}, \mathcal{T}_{caption}).
 \label{generate process}
\end{equation}

\section{Experiments}
The training datasets, training details, experiment setup and more results Please refer the supp. material.
\label{experiments}
% \subsection{Datasets and Training details}
% Transforming the face editing problem from pixel space to semantic space by first editing a mask and then generating an edited face image enhances the controllability of the edits. Additionally, this approach leverages semantic segmentation maps to facilitate the creation of large-scale pairwise training datasets. Notably, we extend a new large-scale dataset consisting of fine-grained amodal segmentation masks for different objects in an input image based on the previous study~\cite{lee2020maskgan}. Then we are able to train the Text-to-Mask model of MuseFace with pairwise data. The overall dataset consists of 30,000 diverse real world face image with semantic segmentation map and a total of 372,767 instance across 18 different semantic classes (e.g., hair, hat, eyeglasses, etc.). We combine the semantic attributes contained in each image using leave-one-out, where an attribute is picked out of it, and the remaining attributes are combined into an intermediate semantic segmentation map until all attributes have been selected. The detailed descriptions $\mathcal{T}_{caption}$ for each image are obtained using image-to-text model.

\begin{table*}[h!]
\centering
\caption{The quantitative results of the edited image. The bold face indicates the best performance and the underline represents the second-best performance. \XSolidBrush indicates the method cannot accomplish the task. Given Talk-to-edit's low success rate with eyeglass editing, it is not possible to generate correspondingly large numbers of images for fair comparisons with FID metric deemed inapplicable.}
\setlength{\tabcolsep}{.35em}
\resizebox{\linewidth}{!}{
\begin{tabular}{c|ccc|cccc|cccc}
\hline \multirow{3}{*}{ Method } & \multicolumn{3}{c|}{ Text-only } & \multicolumn{4}{c|}{ Bounding Box + Text } & \multicolumn{4}{c}{ Attr-wise } \\
\cline { 2 - 12 } & \multirow{2}{*}{ CLIP-Score ($\%$) $\uparrow$ } & \multirow{2}{*}{ FID $\downarrow$ }& \multirow{2}{*}{ ID Sim. $\uparrow$ } & \multirow{2}{*}{ CLIP-Score ($\%$) $\uparrow$ }& \multirow{2}{*}{ Local-FID $\downarrow$ } & \multirow{2}{*}{ ID Sim. $\uparrow$ } & \multirow{2}{*}{ ACC $\uparrow$ } & \multicolumn{2}{c|}{Hair} & \multicolumn{2}{c}{Eyeglasses} \\
\cline{9-12}&&&&&&&&ID Sim. $\uparrow$&\multicolumn{1}{c|}{FID $\downarrow$}&ID Sim. $\uparrow$&FID $\downarrow$\\
\hline 
Instructpix2pix~\cite{brooks2023instructpix2pix} & \underline{24.86} & \textbf{38.92} & 0.47 & \XSolidBrush & \XSolidBrush & \XSolidBrush & \XSolidBrush & \XSolidBrush & \XSolidBrush& \XSolidBrush & \XSolidBrush \\
Talk to edit~\cite{jiang2021talk} & 23.04 & 48.25 & \underline{0.64}&  \XSolidBrush & \XSolidBrush & \XSolidBrush & \XSolidBrush & 0.66 &50.98& 0.58& - \\
Diffae~\cite{preechakul2022diffusion} & \XSolidBrush &\XSolidBrush& \XSolidBrush &  \XSolidBrush& \XSolidBrush & \XSolidBrush & \XSolidBrush & \underline{0.88} &\textbf{42.69}& \underline{0.69}& \underline{58.51} \\
BLDM~\cite{avrahami2023blended} & \XSolidBrush &\XSolidBrush& \XSolidBrush  &23.81& 53.99 & 0.75  & 0.89 & \XSolidBrush & \XSolidBrush& \XSolidBrush & \XSolidBrush \\
SD-inpaint~\cite{rombach2022high} & \XSolidBrush & \XSolidBrush & \XSolidBrush &23.84& \underline{23.03} & \underline{0.83}   & 0.89 & \XSolidBrush & \XSolidBrush& \XSolidBrush & \XSolidBrush \\
SDXL-inpaint~\cite{podell2023sdxl} & \XSolidBrush & \XSolidBrush & \XSolidBrush &23.82& 27.84 & 0.77 & \underline{0.90} & \XSolidBrush & \XSolidBrush& \XSolidBrush & \XSolidBrush \\
% $M^3Face$ & 0.09 & - & - &&& - & - & 0.94 & \XSolidBrush & \XSolidBrush \\
CollabDiff~\cite{huang2023collaborative}& 23.58 & 157.66 & 0.22 &\underline{24.65}& 60.31 & 0.31 & 0.83 & \XSolidBrush & \XSolidBrush& \XSolidBrush & \XSolidBrush \\
\hline Ours & \textbf{25.27} & \underline{47.66} & \textbf{0.82} &\textbf{25.33} & \textbf{16.64} & \textbf{0.86} & \textbf{0.91} & \textbf{0.90} &\underline{49.66}& \textbf{0.74}&\textbf{54.35} \\
\hline
\end{tabular}}
\label{tab:comparison}
% \vspace{-0.3cm}
\end{table*}

\begin{figure}[h]
  \centering
  \includegraphics[width=0.5\textwidth]{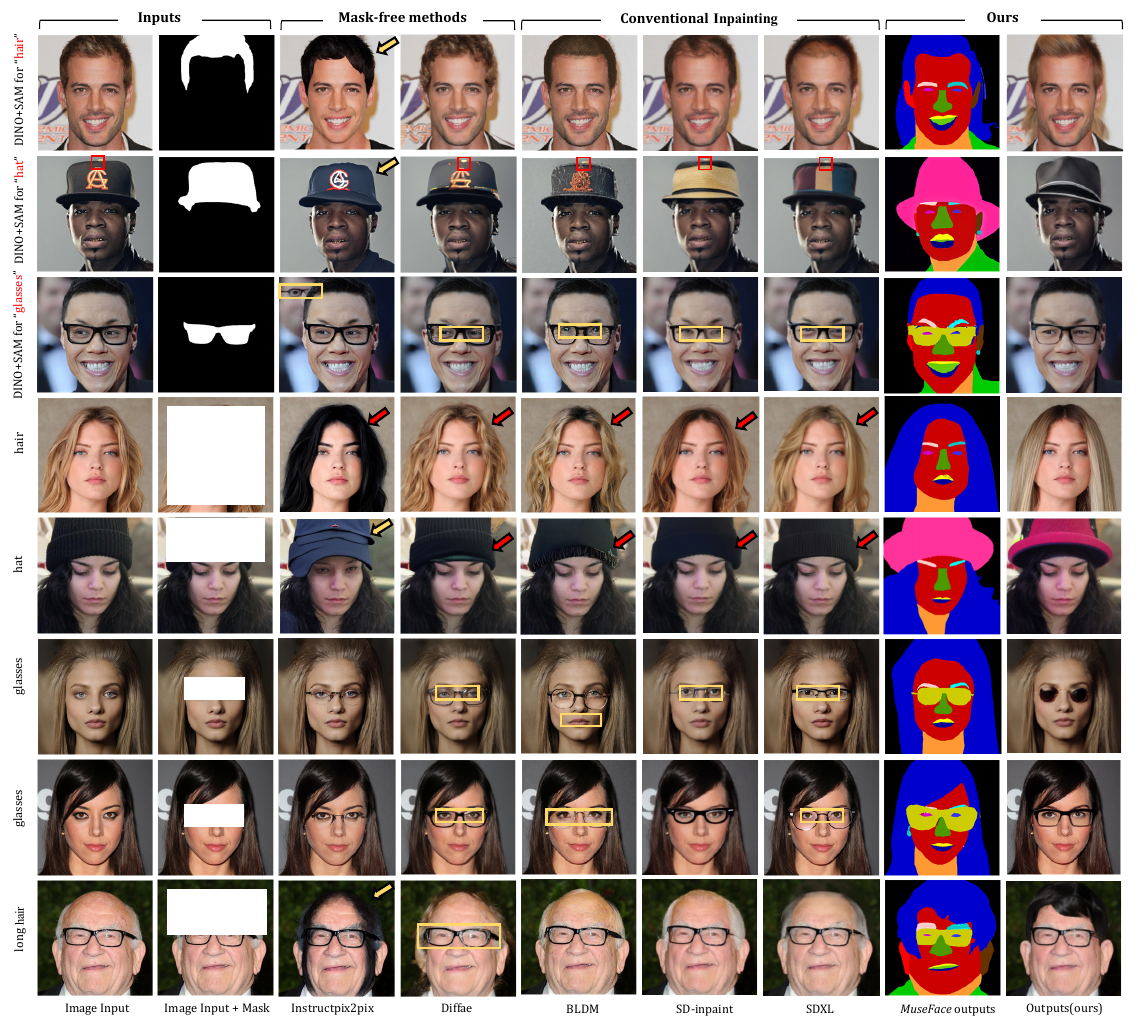}
  \caption{Qualitative comparison of face editing by different methods. MuseFace is capable of handling a wide range of input scenarios with exceptional controllability and flexibility. For a fair comparison, it is recommended to follow the procedure on textual instructions, for instance, offering detailed instructions like ``This man has a medium-length hair.'' instead of single word like ``Hair''. Please zoom in for better visualization and refer supp. material for more results.}
  % \Description{A woman and a girl in white dresses sit in an open car.}
  \label{fig:result_all_compa}
  %\vskip -0.1in
\end{figure}

 % Transforming the face editing problem in pixel space into editing a mask in semantic space before generating an edited face image from the mask not only allows us to improve the controllability of editing, but also allows us to utilize semantic segmentation map to obtain large-scale pairwise training datasets.

\begin{table}[ht]
\centering
\caption{User study. The rating scale ranges from 1 to 5, with higher numbers indicating better performance.}
\setlength{\tabcolsep}{.35em}
\resizebox{\linewidth}{!}{
\begin{tabular}{l|cccc}
\hline
\multirow{2}{*}{\bf{Method}} & \multicolumn{4}{c}{Quality and Controllability} \\
\cline{2-5}
& \bf{ID} $\uparrow$  & \bf{Realism} $\uparrow$ & \bf{Quality} $\uparrow$& \bf{Coherence} $\uparrow$ \\
\hline
Instructpix2pix~\cite{brooks2023instructpix2pix}&3.08&2.56&2.69&2.34\\
Diffae~\cite{preechakul2022diffusion}&3.71&3.32&3.36&3.17\\
BLDM~\cite{avrahami2023blended}&3.54&3.26&2.69&2.96\\
SD-inpaint~\cite{rombach2022high}&3.49&3.34&3.2&2.71\\
\hline
Ours & \textbf{4.20} & \textbf{4.02} & \textbf{4.06}&\textbf{4.06} \\
\hline
\end{tabular} }
\label{tab:user study}
% \vspace{-0.3cm}
\end{table}

\subsection{Quantitative Evaluation}

\textbf{Evaluation Results. }
To verify the validity and generalizability of our framework, we also compare it with other models like text-only driven model and attributes-wise model. Table \ref{tab:comparison} shows the results of the comparison. We observe that MuseFace is capable of a wide range of input scenes and surpass most methods. The MuseFace perform excellently on the mask accuracy and ID similarity, demonstrateing an excellent controllability to editing human face, while delivers high-quality editing tasks (lower Local-Fid). 

We also report the results on the both the realism of generated images and their quality with the inputs prompt by conducting a user study to further evaluate the face editing quality and controllability. We observe that the output of MuseFace have higher ID similarity, realism, quality and coherence. The findings are reported in a quantitative user study in Table \ref{tab:user study}. The findings from the user study demonstrate that our method successfully edits human faces, achieving a high level of quality and controllability, as rated by human observers.

\subsection{Qualitative Results}
We compare the two perspectives of text-guided and coarse-grained mask-guided with text prompt, respectively. Moreover, in order to demonstrate the superiority of our framework more intuitively, we transform the coarse-grained mask into a fine-grained mask by means of SAM~\cite{kirillov2023segany} (due to the limitations of this approach, only coarse-grained masks can be used if it is to add objects such as adding glasses). 

Owing to the Text-to-Mask diffusion model that generates fine-grained masks for guiding face editing, our method is able to perform well on face editing with text input alone. Results are shown in Fig. \ref{fig:result_all_compa}. Our approach has been demonstrated to generate refined semantic masks and enhance the performance of the editing model by providing essential spatial information. We observe that previous inpainting methods have many drawbacks. However, our method successfully addresses and overcomes these issues. For instance, when modifying facial attributes, the mask originates directly from the image, preventing shape adjustments during the inpainting process. This leads to an unsuccessful editing attempt (Red arrows in Fig. \ref{fig:result_all_compa}). Furthermore, if a segment of the editing area is not covered by the mask due to its coarse granularity, that part remains unchanged (Red boxes in Fig. \ref{fig:result_all_compa}). Seriously, when adding object, the image editing changes the non-interested parts due to the inability to convert the bounding box into a refinement mask. This will happen for mask-guided editing models if cannot provide a fine-grained mask. For instance, adding or modifying eyeglasses, because the mask includes eyes fails to keep the eyes as unchanged as possible, the semantics of the mask are unrecognized in the context of editing sunglasses  (Yellow boxes in Fig. \ref{fig:result_all_compa}).  However, our method effectively addresses this challenge. In terms of realism and coherence, our method outperforms compared to mask-free methods (Yellow arrows in Fig. \ref{fig:result_all_compa}). 

\begin{table}[h]
    \caption{Analysis on the effectiveness Mask-aware autoencoder.}
    \label{tab:vae}
    \setlength{\tabcolsep}{.35em}
    \resizebox{\linewidth}{!}{
    \begin{tabular}{lc ccc c cc}
    \toprule
     & & \textbf{Model} & \textbf{Mask-aware} & \textbf{Masked} & & \textbf{PSNR} $\uparrow$ & \textbf{ACC} $\uparrow$  \\
    \midrule
     & & \multirow{3}{*}{{SD VAE~\cite{rombach2022high}}} & None & - & & 30.58 & 0.47 \\
     \multirow{-1}{*}{{\rotatebox[origin=c]{0}{Rescon.}}}& &  & Non-Linear & \XSolidBrush & & 27.27 & 0.65 \\
    \rowcolor{LightCyan}
     \cellcolor{white} &  \cellcolor{white} & \cellcolor{white} & Non-Linear & \Checkmark & & \textbf{37.17} & \textbf{0.76} \\
    \midrule
    & &  & Non-Linear & \Checkmark & & \textbf{15.26} & 0.80 \\
    \rowcolor{LightCyan}
     \cellcolor{white}\multirow{-2}{*}{{\rotatebox[origin=c]{0}{Insertion}}} &  \cellcolor{white} & \cellcolor{white}\multirow{-2}{*}{{Ours}} & None & - & & 14.59 & \textbf{0.90} \\
    \cmidrule{3-8}
    & &  & None & - & & 15.47 & 0.84 \\
    \rowcolor{LightCyan}
     \cellcolor{white}\multirow{-2}{*}{{\rotatebox[origin=c]{0}{Editing}}} &  \cellcolor{white} & \cellcolor{white}\multirow{-2}{*}{{Ours}} & Non-Linear & \Checkmark & & \textbf{16.00} & \textbf{0.89} \\
    \bottomrule
    \end{tabular}}
\vspace{-0.3cm}
\end{table}

\subsection{Additional Study }

\subsubsection{The role of proposed autoencoder}\label{Ablation Study}
We conduct a further study to assess the effectiveness of Mask-aware Autoencoder. Firstly, we report the necessity of the proposed Mask-aware Autoencoder modules in improving the quality of semantic masks reconstruction. The evaluation results of the semantic masks reconstruction setting are illustrated in Table \ref{tab:vae}. It can be observed that the image quality with Mask-aware Autoencoder is superior to the VAE model without Mask-awre Autoencoder in two aspects. For the semantic labeling accuracy, the improvement of our method is more obvious. The results indicate that the Mask-aware Autoencoder is capable of supporting both mask-free and mask-guidance methods. 

We also notice that Mask-aware Autoencoder tends to keep the pixel value constant outside the mask. This property allows for localized editing by changing regions of interest while keeping untargeted regions unchanged in the semantic map. However, this is not conducive to the task of adding objects. For example, when adding eyeglasses in a mask-free manner, the Mask-aware Autoencoder will try to maintain the pixel value as much as possible, which can result in incorrect semantic labels for the region. Therefore, we also compare the effectiveness of the Mask-aware Autoencoder to further verify its role in insertion and editing tasks. In this experiment, MuseFace sets the Mask-aware Autoencoder as a plug-and-play module. The above findings are reported in a quantitative ablation study, shown in Table \ref{tab:vae}. The Mask-aware Autoencoder is employed for editing tasks but not for insertion tasks.

\subsubsection{Cross-ID Face Generation}
\begin{figure}[t]
  \centering
  \includegraphics[width=0.5\textwidth]{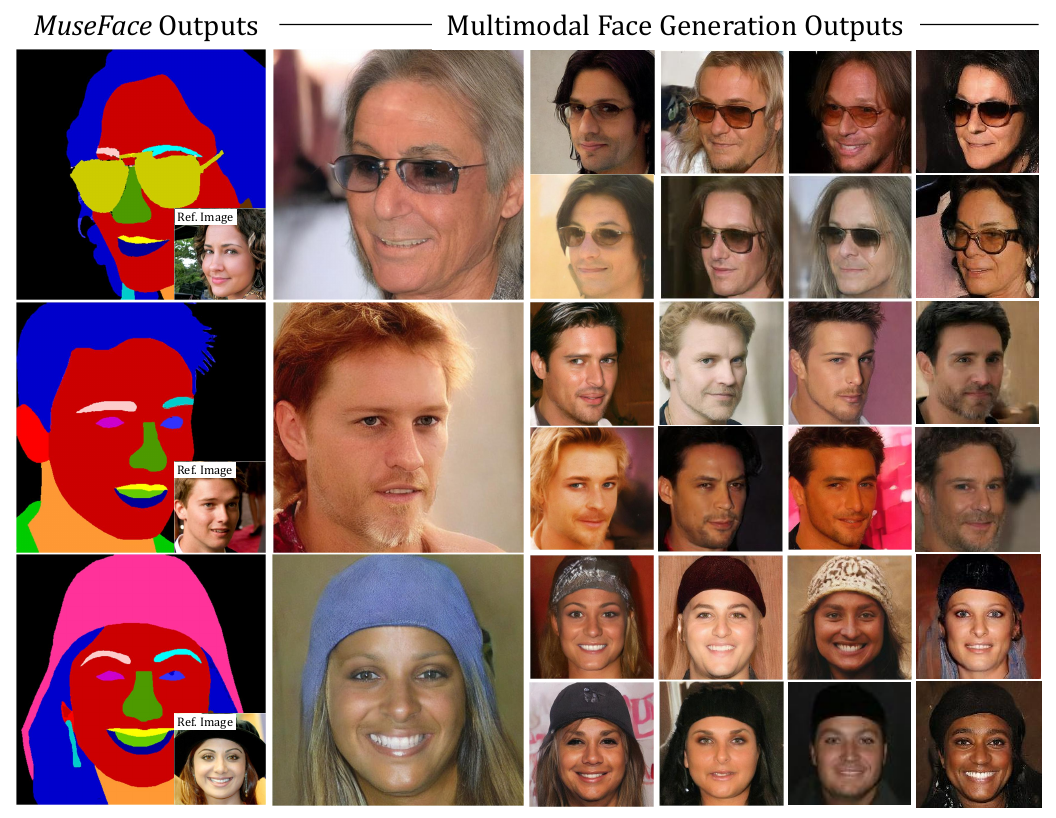}
  \caption{Diverse results of cross-ID generataion. Multimodal face generation is performed using the output from MuseFace, where good consistency is maintained and diversity is also preserved.}
  % \Description{A woman and a girl in white dresses sit in an open car.}
  \label{fig:Face generation}
  % \vskip -0.1in
\end{figure}

We perform the face generation task using the semantic masks generated by MuseFace to further validate the quality of the generated semantic masks. In particular, we use the semantic masks generated by MuseFace for multimodal face generation models~\cite{xia2021tedigan, huang2023collaborative} to generate face images that are consistent with the masks, as shown in Fig. \ref{fig:Face generation}. The experimental results show that the semantic mask edited by MuseFace can also be used for multimodal face generation, providing a base model for subsequent face generation and editing related tasks.

\begin{figure}[h]
  \centering
  \includegraphics[width=0.5\textwidth]{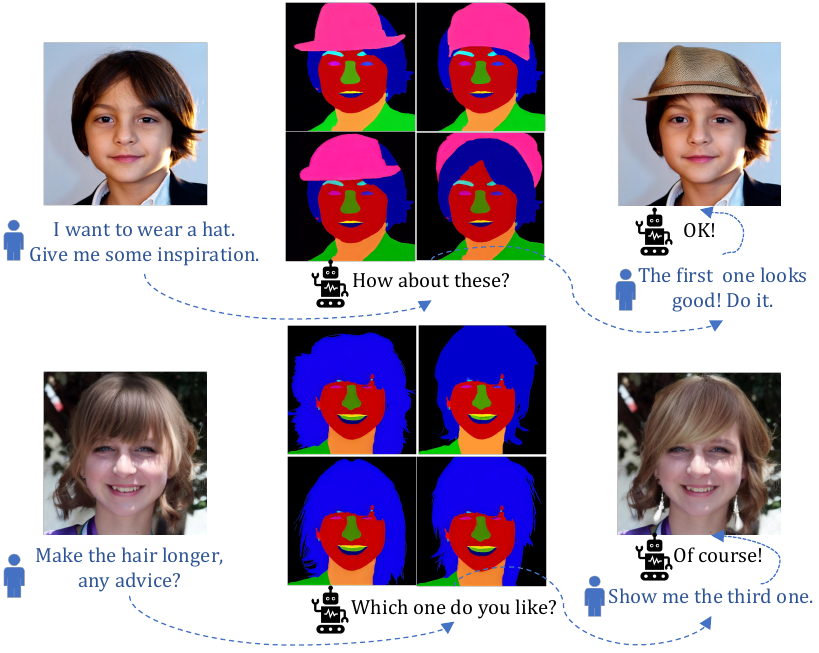}
  \caption{The edited image in the wild. Users can input an image and text, and interact with MuseFace to generate an edited version of the image conveniently.}
  % \Description{A woman and a girl in white dresses sit in an open car.}
  \label{fig:zero-shot}
  % \vskip -0.1in
\end{figure}

\subsubsection{Zero-shot Face Editing}
We test our MuseFace in the image in the wild~\cite{karras2019style}, as shown in Fig. \ref{fig:zero-shot}, which showcases MuseFace's proficiency in zero-shot face editing and the flexibility of MuseFace as an end-to-end editing framework. These findings underscore the model's utility in handling real-world images, illustrating its potential to perform complex editing tasks with remarkable accuracy and efficiency.
% We test our MuseFace in the image in the wild~\cite{karras2019style}, as shown in Figure \ref{fig:zero-shot}, which showcases MuseFace's proficiency in zero-shot face editing and the flexibility of MuseFace as an end-to-end editing framework. These findings underscore the model's utility in handling real-world images, illustrating its potential to perform complex editing tasks with remarkable accuracy and efficiency.

% \input{sections/limitation and discussion}

\section{Conclusion}
\label{conclusion}
We introduce MuseFace, a novel text-driven face editing framework that significantly advances the field of face editing. It uniquely combines a Text-to-Mask diffusion model with a semantic face editing model that utilizes semantic masks to perform precise and detailed edits. The capacity to generate fine-grained masks from textual inputs enhances the precision of edits (controllability) and significantly improves the overall flexibility and diversity of face editing. Our extensive experimental results confirm that MuseFace outperforms existing methods, offering superior fidelity and effectively handling complex editing tasks. By focusing on face editing, we could provide a detailed and thorough analysis of these issues and demonstrate how our method can overcome them, thus filling an important gap. More disscussion refer to supp. material.

\bibliographystyle{IEEEbib}
\bibliography{ref}

\begin{thebibliography}{10}

\bibitem{karras2019style}
T.~Karras, S.~Laine, and T.~Aila,
\newblock ``A style-based generator architecture for generative adversarial networks,''
\newblock in {\em Proceedings of the IEEE/CVF Conference on Computer Vision and Pattern Recognition}, 2019, pp. 4401--4410.

\bibitem{xie2024pointtalkaudiodrivendynamiclip}
Y.~Xie, T.~Feng, X.~Zhang, X.~Luo, Z.~Guo, W.~Yu, H.~Chang, F.~Ma, and F.~R. Yu,
\newblock ``Pointtalk: Audio-driven dynamic lip point cloud for 3d gaussian-based talking head synthesis,''
\newblock {\em arXiv:2412.08504}, 2024.

\bibitem{10126081}
H.~Sun, J.~Ma, Q.~Guo, Q.~Zou, S.~Song, Y.~Lin, and H.~Yu,
\newblock ``Coarse-to-fine task-driven inpainting for geoscience images,''
\newblock {\em IEEE Transactions on Circuits and Systems for Video Technology}, vol. 33, no. 12, pp. 7170--7182, 2023.

\bibitem{Luo2024Code}
X.~Luo, X.~Zhang, Y.~Xie, X.~Tong, W.~Yu, H.~Chang, F.~Ma, and F.~R. Yu,
\newblock ``Codeswap: Symmetrically face swapping based on prior codebook,''
\newblock in {\em Proceedings of the 32nd ACM International Conference on Multimedia}, 2024, p. 6910–6919.

\bibitem{goodfellow2020generative}
I.~Goodfellow, J.~Pouget-Abadie, M.~Mirza, B.~Xu, D.~Warde-Farley, S.~Ozair, A.~Courville, and Y.~Bengio,
\newblock ``Generative adversarial networks,''
\newblock {\em Communications of the ACM}, vol. 63, no. 11, pp. 139--144, 2020.

\bibitem{kingma2013auto}
D.~P Kingma and M.~Welling,
\newblock ``Auto-encoding variational bayes,''
\newblock {\em arXiv:1312.6114}, 2013.

\bibitem{rombach2022high}
R.~Rombach, A.~Blattmann, D.~Lorenz, P.~Esser, and B.~Ommer,
\newblock ``High-resolution image synthesis with latent diffusion models,''
\newblock in {\em Proceedings of the IEEE/CVF Conference on Computer Vision and Pattern Recognition}, 2022, pp. 10684--10695.

\bibitem{ramesh2022hierarchical}
A.~Ramesh, P.~Dhariwal, A.~Nichol, C.~Chu, and M.~Chen,
\newblock ``Hierarchical text-conditional image generation with clip latents,''
\newblock {\em arXiv:2104.08718}, 2022.

\bibitem{saharia2022photorealistic}
C.~Saharia, W.~Chan, S.~Saxena, L.~Li, J.~Whang, E.~L Denton, K.~Ghasemipour, R.~G.~Lopes, B.~K.~Ayan, T.~Salimans, et~al.,
\newblock ``Photorealistic text-to-image diffusion models with deep language understanding,''
\newblock in {\em Proceedings of the 36th International Conference on Neural Information Processing Systems}, 2022.

\bibitem{ho2021classifier}
J.~Ho and T.~Salimans,
\newblock ``Classifier-free diffusion guidance,''
\newblock in {\em NeurIPS 2021 Workshop on Deep Generative Models and Downstream Applications}, 2021.

\bibitem{zhu2017unpaired}
J.~Zhu, T.~Park, P.~Isola, and A.~A Efros,
\newblock ``Unpaired image-to-image translation using cycle-consistent adversarial networks,''
\newblock in {\em Proceedings of the IEEE International Conference on Computer Vision}, 2017, pp. 2223--2232.

\bibitem{durall2021facialgan}
R.~Durall, J.~Jam, D.~Strassel, M.~H. Yap, and J.~Keuper,
\newblock ``Facialgan: Style transfer and attribute manipulation on synthetic faces,''
\newblock {\em arXiv:2110.09425}, 2021.

\bibitem{jiang2021talk}
Y.~Jiang, Z.~Huang, X.~Pan, C.~C. Loy, and Z.~Liu,
\newblock ``Talk-to-edit: Fine-grained facial editing via dialog,''
\newblock in {\em Proceedings of the IEEE/CVF International Conference on Computer Vision}, 2021, pp. 13799--13808.

\bibitem{yue2023chatface}
D.~Yue, Q.~Guo, M.~Ning, J.~Cui, Y.~Zhu, and L.~Yuan,
\newblock ``Chatface: Chat-guided real face editing via diffusion latent space manipulation,''
\newblock {\em arXiv:2305.14742}, 2023.

\bibitem{xia2021tedigan}
W.~Xia, Y.~Yang, J.~Xue, and B.~Wu,
\newblock ``Tedigan: Text-guided diverse face image generation and manipulation,''
\newblock in {\em Proceedings of the IEEE/CVF Conference on Computer Vision and Pattern Recognition}, 2021, pp. 2256--2265.

\bibitem{huang2023collaborative}
Z.~Huang, K.~C. Chan, Y.~Jiang, and Z.~Liu,
\newblock ``Collaborative diffusion for multi-modal face generation and editing,''
\newblock in {\em Proceedings of the IEEE/CVF Conference on Computer Vision and Pattern Recognition}, 2023, pp. 6080--6090.

\bibitem{singh2023smartmask}
J.~Singh, J.~Zhang, Q.~Liu, C.~Smith, Z.~Lin, and L.~Zheng,
\newblock ``Smartmask: Context aware high-fidelity mask generation for fine-grained object insertion and layout control,''
\newblock in {\em Proceedings of the IEEE/CVF Conference on Computer Vision and Pattern Recognition}, 2023, pp. 6497--6506.

\bibitem{brooks2023instructpix2pix}
T.~Brooks, A.~Holynski, and A.~A Efros,
\newblock ``Instructpix2pix: Learning to follow image editing instructions,''
\newblock in {\em Proceedings of the IEEE/CVF Conference on Computer Vision and Pattern Recognition}, 2023, pp. 18392--18402.

\bibitem{qi2019amodal}
L.~Qi, L.~Jiang, S.~Liu, X.~Shen, and J.~Jia,
\newblock ``Amodal instance segmentation with kins dataset,''
\newblock in {\em Proceedings of the IEEE/CVF Conference on Computer Vision and Pattern Recognition}, 2019, pp. 3014--3023.

\bibitem{morelli2023ladi}
D.~Morelli, A.~Baldrati, G.~Cartella, M.~Cornia, M.~Bertini, and R.~Cucchiara,
\newblock ``Ladi-vton: latent diffusion textual-inversion enhanced virtual try-on,''
\newblock in {\em Proceedings of the 31st ACM International Conference on Multimedia}, 2023, pp. 8580--8589.

\bibitem{mou2024t2i}
C.~Mou, X.~Wang, L.~Xie, Y.~Wu, J.~Zhang, Z.~Qi, and Y.~Shan,
\newblock ``T2i-adapter: Learning adapters to dig out more controllable ability for text-to-image diffusion models,''
\newblock in {\em Proceedings of the AAAI Conference on Artificial Intelligence}, 2024, vol.~38, pp. 4296--4304.

\bibitem{zhang2023adding}
L.~Zhang, A.~Rao, and M.~Agrawala,
\newblock ``Adding conditional control to text-to-image diffusion models,''
\newblock in {\em Proceedings of the IEEE International Conference on Computer Vision}, 2023.

\bibitem{lee2020maskgan}
C.~Lee, Z.~Liu, L.~Wu, and P.~Luo,
\newblock ``Maskgan: Towards diverse and interactive facial image manipulation,''
\newblock in {\em Proceedings of the IEEE/CVF Conference on Computer Vision and Pattern Recognition}, 2020, pp. 5549--5558.

\bibitem{preechakul2022diffusion}
K.~Preechakul, N.~Chatthee, S.~Wizadwongsa, and S.~Suwajanakorn,
\newblock ``Diffusion autoencoders: Toward a meaningful and decodable representation,''
\newblock in {\em Proceedings of the IEEE/CVF Conference on Computer Vision and Pattern Recognition}, 2022, pp. 10619--10629.

\bibitem{avrahami2023blended}
O.~Avrahami, O.~Fried, and D.~Lischinski,
\newblock ``Blended latent diffusion,''
\newblock {\em ACM Transactions on Graphics}, vol. 42, no. 4, pp. 1--11, 2023.

\bibitem{podell2023sdxl}
D.~Podell, Z.~English, K.~Lacey, A.~Blattmann, T.~Dockhorn, J.~M{\"u}ller, J.~Penna, and R.~Rombach,
\newblock ``Sdxl: Improving latent diffusion models for high-resolution image synthesis,''
\newblock {\em arXiv:2307.01952}, 2023.

\bibitem{kirillov2023segany}
A.~Kirillov, E.~Mintun, N.~Ravi, H.~Mao, C.~Rolland, L.~Gustafson, T.~Xiao, S.~Whitehead, A.~C. Berg, W.~Lo, P.~Doll{\'a}r, and R.~Girshick,
\newblock ``Segment anything,''
\newblock {\em arXiv:2304.02643}, 2023.

\end{thebibliography}


\begin{thebibliography}{10}

\bibitem{ramesh2022hierarchical}
A.~Ramesh, P.~Dhariwal, A.~Nichol, C.~Chu, and M.~Chen,
\newblock ``Hierarchical text-conditional image generation with clip latents,''
\newblock {\em arXiv:2104.08718}, 2022.

\bibitem{rombach2022high}
R.~Rombach, A.~Blattmann, D.~Lorenz, P.~Esser, and B.~Ommer,
\newblock ``High-resolution image synthesis with latent diffusion models,''
\newblock in {\em Proceedings of the IEEE/CVF Conference on Computer Vision and Pattern Recognition}, 2022, pp. 10684--10695.

\bibitem{saharia2022photorealistic}
C.~Saharia, W.~Chan, S.~Saxena, L.~Li, J.~Whang, E.~L Denton, K.~Ghasemipour, R.~G.~Lopes, B.~K.~Ayan, T.~Salimans, et~al.,
\newblock ``Photorealistic text-to-image diffusion models with deep language understanding,''
\newblock in {\em Proceedings of the 36th International Conference on Neural Information Processing Systems}, 2022.

\bibitem{huang2023composer}
Lianghua Huang, Di~Chen, Yu~Liu, Yujun Shen, Deli Zhao, and Jingren Zhou,
\newblock ``Composer: creative and controllable image synthesis with composable conditions,''
\newblock in {\em Proceedings of the 40th International Conference on Machine Learning}, 2023, pp. 13753--13773.

\bibitem{li2024blip}
Dongxu Li, Junnan Li, and Steven Hoi,
\newblock ``Blip-diffusion: Pre-trained subject representation for controllable text-to-image generation and editing,''
\newblock {\em Advances in Neural Information Processing Systems}, vol. 36, 2024.

\bibitem{ho2021classifier}
J.~Ho and T.~Salimans,
\newblock ``Classifier-free diffusion guidance,''
\newblock in {\em NeurIPS 2021 Workshop on Deep Generative Models and Downstream Applications}, 2021.

\bibitem{nichol2021glide}
Alex Nichol, Prafulla Dhariwal, Aditya Ramesh, Pranav Shyam, Pamela Mishkin, Bob McGrew, Ilya Sutskever, and Mark Chen,
\newblock ``Glide: Towards photorealistic image generation and editing with text-guided diffusion models,''
\newblock {\em arXiv:2112.10741}, 2021.

\bibitem{radford2021learning}
Alec Radford, Jong~Wook Kim, Chris Hallacy, Aditya Ramesh, Gabriel Goh, Sandhini Agarwal, Girish Sastry, Amanda Askell, Pamela Mishkin, Jack Clark, et~al.,
\newblock ``Learning transferable visual models from natural language supervision,''
\newblock in {\em International conference on machine learning}. PMLR, 2021, pp. 8748--8763.

\bibitem{wang2024instantid}
Q.~Wang, X.~Bai, H.~Wang, Z.~Qin, A.~Chen, H.~Li, X.~Tang, and Y.~Hu,
\newblock ``Instantid: Zero-shot identity-preserving generation in seconds,''
\newblock {\em arXiv:2401.07519}, 2024.

\bibitem{mou2024t2i}
C.~Mou, X.~Wang, L.~Xie, Y.~Wu, J.~Zhang, Z.~Qi, and Y.~Shan,
\newblock ``T2i-adapter: Learning adapters to dig out more controllable ability for text-to-image diffusion models,''
\newblock in {\em Proceedings of the AAAI Conference on Artificial Intelligence}, 2024, vol.~38, pp. 4296--4304.

\bibitem{gu2022vector}
Shuyang Gu, Dong Chen, Jianmin Bao, Fang Wen, Bo~Zhang, Dongdong Chen, Lu~Yuan, and Baining Guo,
\newblock ``Vector quantized diffusion model for text-to-image synthesis,''
\newblock in {\em Proceedings of the IEEE/CVF Conference on Computer Vision and Pattern Recognition}, 2022, pp. 10696--10706.

\bibitem{chefer2023attend}
Hila Chefer, Yuval Alaluf, Yael Vinker, Lior Wolf, and Daniel Cohen-Or,
\newblock ``Attend-and-excite: Attention-based semantic guidance for text-to-image diffusion models,''
\newblock {\em ACM Transactions on Graphics (TOG)}, vol. 42, no. 4, pp. 1--10, 2023.

\bibitem{10121479}
Jiadong Liang, Wenjie Pei, and Feng Lu,
\newblock ``Layout-bridging text-to-image synthesis,''
\newblock {\em IEEE Transactions on Circuits and Systems for Video Technology}, vol. 33, no. 12, pp. 7438--7451, 2023.

\bibitem{10126081}
H.~Sun, J.~Ma, Q.~Guo, Q.~Zou, S.~Song, Y.~Lin, and H.~Yu,
\newblock ``Coarse-to-fine task-driven inpainting for geoscience images,''
\newblock {\em IEEE Transactions on Circuits and Systems for Video Technology}, vol. 33, no. 12, pp. 7170--7182, 2023.

\bibitem{zhang2023adding}
L.~Zhang, A.~Rao, and M.~Agrawala,
\newblock ``Adding conditional control to text-to-image diffusion models,''
\newblock in {\em Proceedings of the IEEE International Conference on Computer Vision}, 2023.

\bibitem{meng2021sdedit}
Chenlin Meng, Yutong He, Yang Song, Jiaming Song, Jiajun Wu, Jun-Yan Zhu, and Stefano Ermon,
\newblock ``Sdedit: Guided image synthesis and editing with stochastic differential equations,''
\newblock in {\em International Conference on Learning Representations}, 2021.

\bibitem{avrahami2022blended}
Omri Avrahami, Dani Lischinski, and Ohad Fried,
\newblock ``Blended diffusion for text-driven editing of natural images,''
\newblock in {\em Proceedings of the IEEE/CVF Conference on Computer Vision and Pattern Recognition}, 2022, pp. 18208--18218.

\bibitem{zhou2023maskdiffusion}
Yupeng Zhou, Daquan Zhou, Zuo-Liang Zhu, Yaxing Wang, Qibin Hou, and Jiashi Feng,
\newblock ``Maskdiffusion: Boosting text-to-image consistency with conditional mask,''
\newblock {\em arXiv preprint arXiv:2309.04399}, 2023.

\bibitem{couairon2023diffedit}
Guillaume Couairon, Jakob Verbeek, Holger Schwenk, and Matthieu Cord,
\newblock ``Diffedit: Diffusion-based semantic image editing with mask guidance,''
\newblock in {\em ICLR 2023 (Eleventh International Conference on Learning Representations)}, 2023.

\bibitem{le2024maskdiff}
Minh-Quan Le, Tam~V Nguyen, Trung-Nghia Le, Thanh-Toan Do, Minh~N Do, and Minh-Triet Tran,
\newblock ``Maskdiff: Modeling mask distribution with diffusion probabilistic model for few-shot instance segmentation,''
\newblock in {\em Proceedings of the AAAI Conference on Artificial Intelligence}, 2024, vol.~38, pp. 2874--2881.

\bibitem{singh2023smartmask}
J.~Singh, J.~Zhang, Q.~Liu, C.~Smith, Z.~Lin, and L.~Zheng,
\newblock ``Smartmask: Context aware high-fidelity mask generation for fine-grained object insertion and layout control,''
\newblock in {\em Proceedings of the IEEE/CVF Conference on Computer Vision and Pattern Recognition}, 2023, pp. 6497--6506.

\bibitem{wu2023diffumask}
Weijia Wu, Yuzhong Zhao, Mike~Zheng Shou, Hong Zhou, and Chunhua Shen,
\newblock ``Diffumask: Synthesizing images with pixel-level annotations for semantic segmentation using diffusion models,''
\newblock in {\em Proceedings of the IEEE/CVF International Conference on Computer Vision}, 2023, pp. 1206--1217.

\bibitem{park2023learning}
Minho Park, Jooyeol Yun, Seunghwan Choi, and Jaegul Choo,
\newblock ``Learning to generate semantic layouts for higher text-image correspondence in text-to-image synthesis,''
\newblock in {\em Proceedings of the IEEE/CVF International Conference on Computer Vision}, 2023, pp. 7591--7600.

\bibitem{mofayezi2024m3faceunifiedmultimodalmultilingual}
Mohammadreza Mofayezi, Reza Alipour, Mohammad~Ali Kakavand, and Ehsaneddin Asgari,
\newblock ``M$^3$face: A unified multi-modal multilingual framework for human face generation and editing,'' 2024.

\bibitem{preechakul2022diffusion}
K.~Preechakul, N.~Chatthee, S.~Wizadwongsa, and S.~Suwajanakorn,
\newblock ``Diffusion autoencoders: Toward a meaningful and decodable representation,''
\newblock in {\em Proceedings of the IEEE/CVF Conference on Computer Vision and Pattern Recognition}, 2022, pp. 10619--10629.

\bibitem{huang2023collaborative}
Z.~Huang, K.~C. Chan, Y.~Jiang, and Z.~Liu,
\newblock ``Collaborative diffusion for multi-modal face generation and editing,''
\newblock in {\em Proceedings of the IEEE/CVF Conference on Computer Vision and Pattern Recognition}, 2023, pp. 6080--6090.

\bibitem{xia2021tedigan}
W.~Xia, Y.~Yang, J.~Xue, and B.~Wu,
\newblock ``Tedigan: Text-guided diverse face image generation and manipulation,''
\newblock in {\em Proceedings of the IEEE/CVF Conference on Computer Vision and Pattern Recognition}, 2021, pp. 2256--2265.

\bibitem{ding2023diffusionrig}
Zheng Ding, Xuaner Zhang, Zhihao Xia, Lars Jebe, Zhuowen Tu, and Xiuming Zhang,
\newblock ``Diffusionrig: Learning personalized priors for facial appearance editing,''
\newblock in {\em Proceedings of the IEEE/CVF Conference on Computer Vision and Pattern Recognition}, 2023, pp. 12736--12746.

\bibitem{xu2024personalized}
Jianjin Xu, Saman Motamed, Praneetha Vaddamanu, Chen~Henry Wu, Christian Haene, Jean-Charles Bazin, and Fernando De~la Torre,
\newblock ``Personalized face inpainting with diffusion models by parallel visual attention,''
\newblock in {\em Proceedings of the IEEE/CVF Winter Conference on Applications of Computer Vision}, 2024, pp. 5432--5442.

\bibitem{jiang2021talk}
Y.~Jiang, Z.~Huang, X.~Pan, C.~C. Loy, and Z.~Liu,
\newblock ``Talk-to-edit: Fine-grained facial editing via dialog,''
\newblock in {\em Proceedings of the IEEE/CVF International Conference on Computer Vision}, 2021, pp. 13799--13808.

\bibitem{yue2023chatface}
D.~Yue, Q.~Guo, M.~Ning, J.~Cui, Y.~Zhu, and L.~Yuan,
\newblock ``Chatface: Chat-guided real face editing via diffusion latent space manipulation,''
\newblock {\em arXiv:2305.14742}, 2023.

\bibitem{gu2019mask}
Shuyang Gu, Jianmin Bao, Hao Yang, Dong Chen, Fang Wen, and Lu~Yuan,
\newblock ``Mask-guided portrait editing with conditional gans,''
\newblock in {\em Proceedings of the IEEE/CVF conference on computer vision and pattern recognition}, 2019, pp. 3436--3445.

\bibitem{lee2020maskgan}
C.~Lee, Z.~Liu, L.~Wu, and P.~Luo,
\newblock ``Maskgan: Towards diverse and interactive facial image manipulation,''
\newblock in {\em Proceedings of the IEEE/CVF Conference on Computer Vision and Pattern Recognition}, 2020, pp. 5549--5558.

\bibitem{morelli2023ladi}
D.~Morelli, A.~Baldrati, G.~Cartella, M.~Cornia, M.~Bertini, and R.~Cucchiara,
\newblock ``Ladi-vton: latent diffusion textual-inversion enhanced virtual try-on,''
\newblock in {\em Proceedings of the 31st ACM International Conference on Multimedia}, 2023, pp. 8580--8589.

\bibitem{loshchilov2018decoupled}
Ilya Loshchilov and Frank Hutter,
\newblock ``Decoupled weight decay regularization,''
\newblock in {\em International Conference on Learning Representations}, 2018.

\bibitem{von-platen-etal-2022-diffusers}
Patrick von Platen, Suraj Patil, Anton Lozhkov, Pedro Cuenca, Nathan Lambert, Kashif Rasul, Mishig Davaadorj, Dhruv Nair, Sayak Paul, William Berman, Yiyi Xu, Steven Liu, and Thomas Wolf,
\newblock ``Diffusers: State-of-the-art diffusion models,'' \url{https://github.com/huggingface/diffusers}, 2022.

\bibitem{brooks2023instructpix2pix}
T.~Brooks, A.~Holynski, and A.~A Efros,
\newblock ``Instructpix2pix: Learning to follow image editing instructions,''
\newblock in {\em Proceedings of the IEEE/CVF Conference on Computer Vision and Pattern Recognition}, 2023, pp. 18392--18402.

\bibitem{avrahami2023blended}
O.~Avrahami, O.~Fried, and D.~Lischinski,
\newblock ``Blended latent diffusion,''
\newblock {\em ACM Transactions on Graphics}, vol. 42, no. 4, pp. 1--11, 2023.

\bibitem{podell2023sdxl}
D.~Podell, Z.~English, K.~Lacey, A.~Blattmann, T.~Dockhorn, J.~M{\"u}ller, J.~Penna, and R.~Rombach,
\newblock ``Sdxl: Improving latent diffusion models for high-resolution image synthesis,''
\newblock {\em arXiv:2307.01952}, 2023.

\bibitem{heusel2017gans}
Martin Heusel, Hubert Ramsauer, Thomas Unterthiner, Bernhard Nessler, and Sepp Hochreiter,
\newblock ``Gans trained by a two time-scale update rule converge to a local nash equilibrium,''
\newblock {\em Advances in neural information processing systems}, vol. 30, 2017.

\bibitem{Seitzer2020FID}
Maximilian Seitzer,
\newblock ``{pytorch-fid: FID Score for PyTorch},'' \url{https://github.com/mseitzer/pytorch-fid}, August 2020,
\newblock Version 0.3.0.

\bibitem{hessel2021clipscore}
Jack Hessel, Ari Holtzman, Maxwell Forbes, Ronan~Le Bras, and Yejin Choi,
\newblock ``Clipscore: A reference-free evaluation metric for image captioning,''
\newblock {\em arXiv:2104.08718}, 2021.

\bibitem{wang2018cosface}
Hao Wang, Yitong Wang, Zheng Zhou, Xing Ji, Dihong Gong, Jingchao Zhou, Zhifeng Li, and Wei Liu,
\newblock ``Cosface: Large margin cosine loss for deep face recognition,''
\newblock in {\em Proceedings of the IEEE conference on computer vision and pattern recognition}, 2018, pp. 5265--5274.

\bibitem{DBLP:journals/corr/abs-2312-11396}
Qi~Mao, Lan Chen, Yuchao Gu, Zhen Fang, and Mike~Zheng Shou,
\newblock ``Mag-edit: Localized image editing in complex scenarios via mask-based attention-adjusted guidance,''
\newblock {\em CoRR}, vol. abs/2312.11396, 2023.

\bibitem{DBLP:conf/aaai/YuLFM024}
Zihao Yu, Haoyang Li, Fangcheng Fu, Xupeng Miao, and Bin Cui,
\newblock ``Accelerating text-to-image editing via cache-enabled sparse diffusion inference,''
\newblock in {\em Thirty-Eighth {AAAI} Conference on Artificial Intelligence, {AAAI} 2024, Thirty-Sixth Conference on Innovative Applications of Artificial Intelligence, {IAAI} 2024, Fourteenth Symposium on Educational Advances in Artificial Intelligence, {EAAI} 2014, February 20-27, 2024, Vancouver, Canada}, Michael~J. Wooldridge, Jennifer~G. Dy, and Sriraam Natarajan, Eds. 2024, pp. 16605--16613, {AAAI} Press.

\end{thebibliography}

\end{document}

\title{Supplementary Materials: MuseFace: Text-driven Face Editing via Diffusion-based Mask Generation Approach}

% \author{Anonymous ICME submission}

\maketitle

% \section{Overview}
% We provide a detailed supplementary materials for our work, which explains and complements our methodology in more detail. First of all, we elaborate on the network structure of $MuseFace$. Then, we show additional results.

\section{Related work}\label{related work}
\textbf{Text-to-image Diffusion Models. }In recent years, text-to-image diffusion models~\cite{ramesh2022hierarchical, rombach2022high, saharia2022photorealistic, huang2023composer, li2024blip} have attracted significant attention in the field of AIGC for their ability to produce high-quality, photorealistic images from textual descriptions. The widespread strategy involves denoising in the latent space and incorporating the text condition into the denoising process through cross-attention modules. One pioneering work in this area is GLIDE by OpenAI, which uses classifier-free guidance~\cite{ho2021classifier} to improve the photorealism and relevance of generated images to their text descriptions~\cite{nichol2021glide}. Based on the idea of alignment, CLIP~\cite{radford2021learning}, a vision-language model 
which aligns visual and textual inputs in a shared embedding space, is commonly employed as a text encoder for the text-to-image diffusion model~\cite{wang2024instantid, mou2024t2i, gu2022vector, chefer2023attend}. Although they achieve promising synthesis quality, the text prompt can not provide the synthesis results with reliable structural guidance~\cite{mou2024t2i}. Therefore, several works~\cite{10121479, 10126081} explore the potential of introducing additional conditions into the text-to-image diffusion model to guide the generation process. For example, ControlNet~\cite{zhang2023adding} introduces spatial conditioning controls into large, pretrained text-to-image diffusion models. This enhancement enables more precise and controllable results, thus significantly advancing this field of exploration.

\textbf{Mask Generation. }
Masks provide one of the common and important forms of spatial conditioning, simplifying the task by treating it as a conditional inpainting task~\cite{meng2021sdedit, avrahami2022blended,zhou2023maskdiffusion}. The degree of refinement in the mask directly enhances the accuracy with which text-to-image diffusion models synthesize natural images. However, it is extremely challenging to get users to directly provide refined masks for accurate generation. To address this challenge, Diffedit~\cite{couairon2023diffedit} automatically generates a mask that highlights regions of the input image requiring editing, which is achieved by contrasting predictions from a diffusion model conditioned on different text prompts. At the same time, MaskDiff~\cite{le2024maskdiff} gets the mask of the corresponding object in natural image from the idea of instance segmentation models. More freely, SmartMask~\cite{singh2023smartmask} allows any novice user to generate precise object masks for fine-grained object insertion with better background preservation. Recently, some works have focused on generating semantic segmentation which is a specialized masks utilizing different pixel values to label objects~\cite{wu2023diffumask, park2023learning}. In particular, M$^3$Face~\cite{mofayezi2024m3faceunifiedmultimodalmultilingual} is capable of generating semantic segmentation of the face for controllable face generation, but fall short in editing because it can not generate the edited semantic map for face editing. In contrast previous studies, MuseFace can generate edited semantic segmentation for human face editing in an end-to-end manner.

\textbf{Human face editing. }
Human face editing~\cite{mofayezi2024m3faceunifiedmultimodalmultilingual, preechakul2022diffusion, huang2023collaborative,xia2021tedigan,ding2023diffusionrig, xu2024personalized} aims to modify some of the attributes of the face (e.g., long hair to short hair), which has a wide range of application scenarios and commercial value. GAN and diffusion-based generative models perform well on this task. For instance, some studies attempt to implement face editing using only text prompt, such as Talk-to-edit~\cite{jiang2021talk} and ChatFace~\cite{yue2023chatface}. Furthermore, some methods focus on image synthesis conditioned on a variety of user-provided guidance for interactive and controllable face editing, e.g., mask~\cite{gu2019mask}, or semantic masks~\cite{mofayezi2024m3faceunifiedmultimodalmultilingual}. While these methods benefit from different modalities to achieve stability and quality, they often pose challenges to user experience due to the intricate process of creating inputs like segmentation maps. In contrast, we introduce MuseFace, a unified text-driven face editing framework, where the user inputs a reference image and a text describing the part they want to edit. Our framework first predicts the shape of the edited mask by the Text-to-Mask diffusion model, and then generates the edited natural image with guidance from the edited semantic mask.

\section{Model Architecture}
\subsection{Mask-aware Autoencoder}
\begin{figure}[h]
  \centering
  \includegraphics[width=0.5\textwidth]{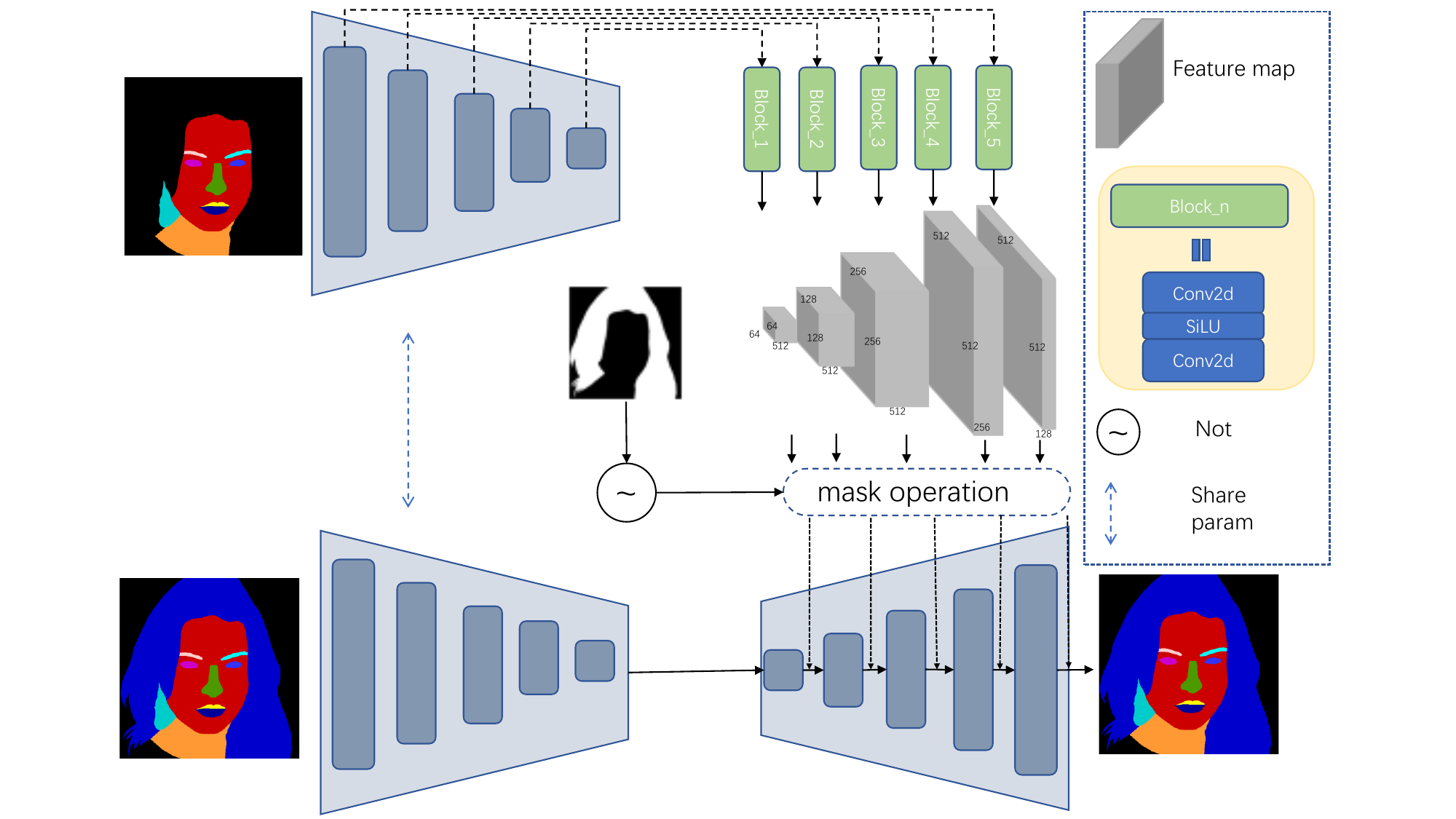}
  \caption{Overview of the proposed autoencoder with Mask-aware modules.}
  % \Description{A woman and a girl in white dresses sit in an open car.}
  \label{fig:autoencoder}
  % \vskip -0.1in
\end{figure}

In this study, the encoding-decoding framework is augmented with an auxiliary branch. Specifically, each feature layer from the encoder of auxiliary branch is aggregated with its corresponding layer in the decoder of the main branch through convolution modules. An overview of Mask-aware Autoencoder is illustrated in Figure \ref{fig:autoencoder}.

\section{Experiments}
\subsection{Datasets and Training details}
Transforming the face editing problem from pixel space to semantic space by first editing a mask and then generating an edited face image enhances the controllability of the edits. Additionally, this approach leverages semantic segmentation maps to facilitate the creation of large-scale pairwise training datasets. Notably, we extend a new large-scale dataset consisting of fine-grained amodal segmentation masks for different objects in an input image based on the previous study~\cite{lee2020maskgan}. Then we are able to train the Text-to-Mask model of MuseFace with pairwise data. The overall dataset consists of 30,000 diverse real world face image with semantic segmentation map and a total of 372,767 instance across 18 different semantic classes (e.g., hair, hat, eyeglasses, etc.). We combine the semantic attributes contained in each image using leave-one-out, where an attribute is picked out of it, and the remaining attributes are combined into an intermediate semantic segmentation map until all attributes have been selected. The detailed descriptions $\mathcal{T}_{caption}$ for each image are obtained using image-to-text model.

We first train the Mask-aware Autoencoder $\mathcal{A}$ for 30k steps with batch size of 2 in NVIDIA A100-40G GPUs. The skip connection implementation details and hyperparameters apply as same as ladi-vton~\cite{morelli2023ladi}. During training, the encoder $\mathcal{E}$ and decoder $\mathcal{D}$ are frozen. Then, we freeze the weights of Mask-aware Autoencoder and train the U-Net of diffusion face mask generation model. In order to leverage the rich generalizable prior of text-to-image diffusion models, we use the weights from publicly available Stable-Diffusion\footnote{\url{https://huggingface.co/runwayml/stable-diffusion-v1-5}} to initialize the weights of the MuseFace U-Net models. We modify the architecture of the U-Net model to also condition the output mask editing on the segmentation map $S_{n}$. We then train the U-Net for a total of 100k steps with batch size of 48. Such training process is full parameter training with learning rate 1e-5 with 500 warm-up steps using a linear schedule. AdamW~\cite{loshchilov2018decoupled} is used as the optimizer with $\beta_1=0.9,\beta_2=0.999$ and weight decay equal to 1e-2. Then, we use this as a base model and only modifications can be accomplished at this point without the ability to add objects due to the limitations of the training data. Next, we freeze all layers except the $conv_{in}$ layer, fine-tune the base model with same setting as above, and then implement object-specific additions such as eyeglasses. To perform face editing, we also train a semantic-aware face editing model~\cite{zhang2023adding, von-platen-etal-2022-diffusers} with SD-1.5 backbone for performing accurate face editing with fine-grained mask outputs.

\subsection{Setup}
\textbf{Baselines. }We first evaluate our model against prior works that utilize text-only inputs for face editing. These include Talk-to-edit~\cite{jiang2021talk}, Instructpix2pix~\cite{brooks2023instructpix2pix}, and text-only CollabDiff~\cite{huang2023collaborative}. Next, we assess the performance of our model in scenarios involving multimodal inputs, comparing it with BLDM~\cite{avrahami2023blended}, SD-inpaint~\cite{rombach2022high}, SDXL-inpaint~\cite{podell2023sdxl}, and CollabDiff~\cite{huang2023collaborative}. Additionally, we explore the effectiveness of attributes-wise editing by comparing our model to Talk-to-edit~\cite{jiang2021talk} and diffae~\cite{preechakul2022diffusion}.

\textbf{Evaluation Metrics. }For text-only driven face editing tasks, we report the results using Local-Fr\'{e}chet Inception Distance (Local-FID)~\cite{heusel2017gans, Seitzer2020FID} which indicates to the effects generated by local editing, CLIP-Score~\cite{hessel2021clipscore} which measures the alignment of the generated images and text, and ID similarity~\cite{wang2018cosface} measuring the consistency of the identity between the two images. Unlike the text-only driven face editing task, we additionally add mask accuracy to indicate controllability with mask accuracy~\cite{huang2023collaborative} in the multimodal face editing task. For each output image, we predict the segmentation mask using the face parsing network provided by~\cite{lee2020maskgan}. Mask accuracy is the pixel-wise accuracy against the ground-truth segmentation. A higher average accuracy indicates better consistency between the output. We also use ID similarity and FID to measure the performance of our model for a single attribute.

\subsection{More Experiments and Results}
\subsubsection{User Control}
\begin{figure}[h!]
  \centering
  \includegraphics[width=0.5\textwidth]{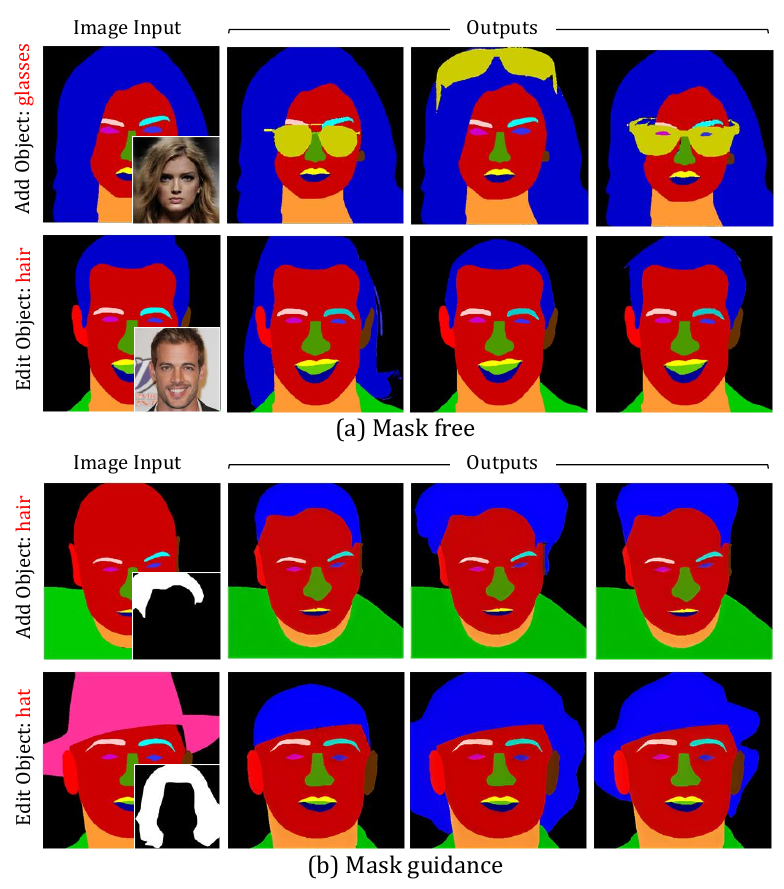}
  \caption{Diverse user controls and suggestions. MuseFace is able to generate masks in both mask-free and mask-guided manner. It also provides a variety of generation results.}
  % \Description{A woman and a girl in white dresses sit in an open car.}
  \label{fig:result}
  % \vskip -0.1in
\end{figure}

% 这里插一个图，掩码的对比
A key advantage of MuseFace is the ability to generate high-quality masks for target object in a controllable manner and edit the face based on the masks with high fidelity. We evaluate the performance of MuseFace in terms of user control. As shown in Fig. \ref{fig:result}, MuseFace allows the user to control the output object mask in two main ways to perform edit and add object. Importantly, MuseFace allows users to edit and add objects by providing only text, shown as Fig. \ref{fig:result} (a). By mask-free manner, MuseFace provides editing inspiration and greatly reduces the difficulty of editing while guaranteeing a controlled generation of edits. Additionally, the user can directly take advantage of MuseFace to generate a variety of masks for a given location by supplying a coarse or fine grained mask to specify the edit location, shown as Fig. \ref{fig:result} (b). Such user-provided masks do not provide any information for editing anything other than location, such as shape, ensuring the diversity of the MuseFace. In this way, MuseFace is able to handle some hard cases and produce great results.

\subsubsection{More Results}

\begin{figure*}[]
  \centering
  \includegraphics[width=\textwidth]{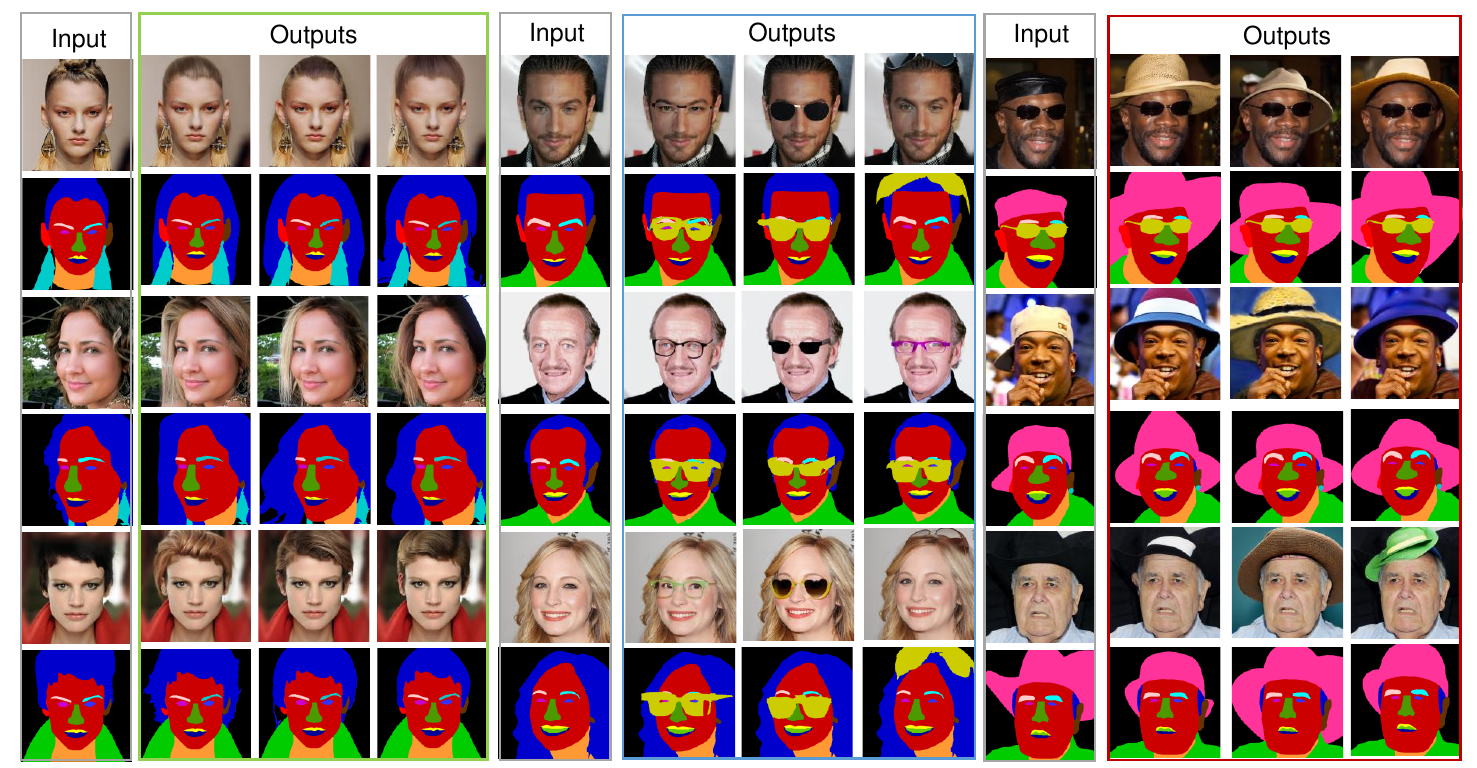}
  \caption{More qualitative results of face editing by our model. Please zoom in for better visualization.}
  % \Description{A woman and a girl in white dresses sit in an open car.}
  \label{fig:more result}
  %\vskip -0.1in
\end{figure*}

More results of MuseFace are shown in Fig. \ref{fig:more result}, demonstrating the superiority of our method in terms of performance.

\section{DISCUSSION}\label{discussion}
Inspiringly, we demonstrate that MuseFace can serve as a foundational model for downstream tasks, such as Cross-ID Face Generation. The masks generated by MuseFace guide the multimodal face generation model to maintain excellent consistency and diversity in the generated images. Moreover, the implicit conditional control can lead to a gap in generalization ability when guiding image generation. When releasing the pre-trained weights, we will further expand the dataset for training the conditional network to enhance the generalization ability of multimodal face editing model.

What's more, there are many exiting methods which also generate masks from images and then use these masks as inputs for inpainting. However, our proposed method innovatively improves fine-grained object editing performance, which is different from many existing methods. \textbf{a)}: Impressive works such as MAG-Edit \cite{DBLP:journals/corr/abs-2312-11396} and FISEdit \cite{DBLP:conf/aaai/YuLFM024} rely on refinement masks provided by users for editing. However, creating refinement masks increases the editing cost for users and introduces variability between different users. \textbf{b)}: Based on this, some methods like Diffedit and Smartmask attempt to guide editing using generated masks. Experiments have shown that this approach performs well for coarse-grained object editing, such as for humans, horses, cats, etc. However, it has shortcomings in editing fine-grained objects, such as specific attributes of coarse-grained objects. The main reason is that these generated masks have weak edge awareness, leading to poor quality masks for fine-grained objects. Extensive experiments validate MuseFace's ability of fine-grained editing.\\
\bibliographystyle{IEEEbib}
\bibliography{ref}